\documentclass{IEEEtaes}

\usepackage{color,array,amsthm}
\usepackage{graphicx}

\usepackage{subcaption}
\usepackage{booktabs}
\usepackage{amsmath}
\usepackage{algorithm}
\usepackage{algpseudocode}

\jvol{XX}
\jnum{XX}
\jmonth{XXXXX}
\paper{1234567}
\pubyear{2022}
\doiinfo{TAES.2022.Doi Number}

\begin{document}

\title{SpY: A Context-Based Approach to Spacecraft Component Detection}

\author{TRUPTI MAHENDRAKAR}
\member{Member, IEEE}
\affil{Florida Institute of Technology, Melbourne, FL 32901, USA} 

\author{RYAN T. WHITE}
\affil{Florida Institute of Technology, Melbourne, FL 32901, USA} 

\author{MADHUR TIWARI}
\affil{Florida Institute of Technology, Melbourne, FL 32901, USA}

\receiveddate{Manuscript received XXXXX 00, 0000; revised XXXXX 00, 0000; accepted XXXXX 00, 0000.\\
This work was supported in part by the NVIDIA Applied Research Accelerator Program.}

\corresp{{\itshape (Corresponding author: T. Mahendrakar)}.}


\authoraddress{Trupti Mahendrakar (e-mail: \href{mailto:tmahendrakar2020@my.fit.edu}{tmahendrakar2020@my.fit.edu}) and Madhur Tiwari (e-mail: \href{mailto:mtiwari@fit.edu}{mtiwari@fit.edu}) are with the Department of Aerospace, Physics and Space Sciences and Ryan T. White (e-mail: \href{mailto:rwhite@fit.edu}{rwhite@fit.edu}) is with the Department of Mathematics and Systems Engineering at Florida Institute of Technology, Melbourne, FL 32901 USA.}

\editor{The datasets and SpY algorithm will be made publicly available.}

\markboth{MAHENDRAKAR ET AL.}{SPY}
\maketitle

\begin{abstract}
This paper focuses on autonomously characterizing components such as solar panels, body panels, antennas, and thrusters of an unknown resident space object (RSO) using camera feed to aid autonomous on-orbit servicing (OOS) and active debris removal. Significant research has been conducted in this area using convolutional neural networks (CNNs). While CNNs are powerful at learning patterns and performing object detection, they struggle with missed detections and misclassifications in environments different from the training data, making them unreliable for safety in high-stakes missions like OOS. Additionally, failures exhibited by CNNs are often easily rectifiable by humans using commonsense reasoning and contextual knowledge. Embedding such reasoning in an object detector could improve detection accuracy. To validate this hypothesis, this paper presents an end-to-end object detector called SpaceYOLOv2 (SpY), which leverages the generalizability of CNNs while incorporating contextual knowledge using traditional computer vision techniques. SpY consists of two main components: a shape detector and the SpaceYOLO classifier (SYC). The shape detector uses CNNs to detect primitive shapes of RSOs and SYC associates these shapes with contextual knowledge, such as color and texture, to classify them as spacecraft components or ``unknown'' if the detected shape is uncertain. SpY’s modular architecture allows customizable usage of contextual knowledge to improve detection performance, or SYC as a secondary fail-safe classifier with an existing spacecraft component detector. Performance evaluations on hardware-in-the-loop images of a mock-up spacecraft demonstrate that SpY is accurate and an ensemble of SpY with a previously used CNN spacecraft component detector improved the performance by 23.4\% in recall, demonstrating enhanced safety for CNNs in vision-based navigation tasks.

\end{abstract}


\section{INTRODUCTION}
W{\scshape ith} the rapid proliferation of space debris containing retired and defunct satellites, autonomous on-orbit servicing (OOS) and active debris removal (ADR) have gained significant interest. Many of the satellites requiring OOS and ADR are large, unknown, and non-cooperative by nature. They are not equipped with capture interfaces, may be tumbling, and may have endured structural damage. Despite efforts in the literature, this remains an unsolved problem.

To tackle this issue, our previous work focused on autonomously characterizing these unknown targets using convolutional neural network (CNN) based object detectors \cite{Mahendrakar2024, Mahendrakar2021, Mahendrakar2022} to identify potential capture points and keep-out zones. Due to the lack of real spacecraft imagery and to replicate a real-life unknown resident space object (RSO) scenario, the CNNs were trained on a synthetic dataset of random spacecraft images and tested on a never-before-seen hardware-in-the-loop images of a mock-up spacecraft. The components detected include solar panels, antennas, body panels, and thrusters. The 3D positions of these components were resolved using several camera observers with CNN detections are fed into an artificial potential field guidance algorithm \cite{Cutler2022, mahendrakar2023autonomous} to enable safe RPO trajectories for the chaser spacecraft.

Laboratory experimental test results of this concept, discussed in \cite{mahendrakar2023autonomous} revealed that the success of this type of mission is highly dependent on the performance of the CNN object detector. CNN-based object detectors rely heavily on the similarity and patterns seen in the training dataset. This reliance often results in missed detections or misclassifications in real-world scenarios, where varying environmental conditions such as lighting and viewing angles and dissimilarity in training and testing dataset are prevalent. Both missed detections and misclassifications pose a safety threat in using CNNs. Humans use context-based reasoning to detect spacecraft components (e.g., recognizing a long, protruding, rectangular object pointed towards the sun as a solar panel).

To encode this untapped human reason into an autonomous system, this work presents SpaceYOLOv2 (SpY), an end-to-end, human-directed, context-based object detector. This work builds upon SpaceYOLO \cite{SpaceYOLO}, which conducted a survey of aerospace professionals revealing that geometry, texture, and color are the top criteria for identifying spacecraft components by humans. SpaceYOLO demonstrated the proof-of-concept feasibility of using the YOLOv5 CNN to detect primitive shapes such as circles and rectangles in spacecraft images and then classifying them using texture features. However, it lacked a complete end-to-end object detection and performance levels required for practical use. SpY includes a drastically more robust shape detector and a spacecraft component classifier (SYC) based on shape, color, and texture feature extraction methods from traditional computer vision. The shape detector and SYC work together to incorporate contextual reasoning for component detection. SpY is much more robust, achieves competitive accuracies, and has several fault tolerance mechanisms for spacecraft component detection.

The main contributions of this work include:
\begin{enumerate}
    \item An end-to-end object detection pipeline that incorporates contextual knowledge.
    \item A new tool for creating a shape detector training dataset (explained in Section III).
    \item Expanded SYC to incorporate entropy-based (texture) and color-based classifications.
    \item Made SYC modular to use as a secondary classifier with any spacecraft component object detector.
\end{enumerate}


The rest of the paper is structured as follows: Section II discusses the background, including related missions and CNN-based computer vision for on-orbit applications. Section III provides an overview of the methods evaluated and the datasets used in this study. Section IV discusses the SpY pipeline. Section V includes metrics used in this study and presents the results and analysis. Finally, the conclusion is given in Section VI.


\section{BACKGROUND}

\subsection{Related Missions}

The concepts of ADR and OOS have been integral to the space industry since its inception. Manned OOS missions, such as those performed by the space shuttle, have demonstrated the benefits of repairing and extending the life of satellites like the Hubble Space Telescope, Palapa B, and Westar VI \cite{Goodman2006}. Subsequently, robotic OOS missions—starting with ETS VII by JAXA in 1997 \cite{Yoshida2004}, and followed by XSS-10 \cite{Davis2004}, XSS-11 \cite{afrl_xss-11_2011}, ANGELS \cite{afrl_automated_2014}, and Orbital Express \cite{Kennedy2008} by NASA, DARPA, and AFRL—have showcased OOS capabilities with cooperative spacecraft. These spacecraft maintained stable attitudes, were equipped with load-bearing capture interfaces for robotic manipulators, and featured visible fiducial markings for relative navigation.
In 2020 and 2021, Northrop Grumman’s MEV-1 and MEV-2 \cite{Pyrak2022} demonstrated the first commercial OOS with GEO satellites IS-901 and IS-10-02. Despite IS-901 being non-cooperative and tumbling, the presence of distinct apogee kick motors and launch adapter rings (common GEO spacecraft features) facilitated docking.

However, rendezvous and proximity operations (RPO) around unknown spacecraft without these distinct docking features remain challenging. SpY aims to address this by using contextual descriptions to classify features as potential docking or keep-out zones, facilitating safe docking and capture. For example, conical features like apogee kick motors or flat body panels would be suitable for docking, while thin, fragile solar panels should be avoided.

\subsection{CNN for RPO and OOS Tasks}

Over the past 15 years, CNNs have revolutionized computer vision. The development of large datasets has led to more efficient and accurate algorithms. Computing resources have become cheaper and faster, particularly for highly parallelized CNNs accelerated by graphics processing units (GPUs). Recent advancements in low size, weight, and power (SWaP) computers equipped with small GPUs or field-programmable gate arrays (FPGAs) \cite{Ekblad2023} have enabled the deployment of CNNs on spacecraft.

Numerous studies propose CNNs for in-space use. We focus on object detection for spacecraft components (solar arrays, antennas, thrusters, and satellite bodies), where the goal is to predict a bounding box around each component and classify what is in 2D image frames. However, much research has been done on other vision tasks like pose estimation and instance segmentation. Notable are participating works \cite{Black2021, Hogan2021, Rondao2023, Garcia2021, Zhou2022} in ESA’s spacecraft pose estimation challenges for non-cooperative spacecraft based on the SPEED \cite{Speed} and SPEED+ \cite{Park2022} datasets. Further, tasks are sometimes combined: numerous studies use object detection to find a region of interest (ROI) containing the entire target spacecraft in a camera frame, which can be extracted and subjected to downstream analysis. For example, past research has used YOLOv3 to detect CubeSats on a Raspberry Pi \cite{Aarestad2020}, U-Net for spacecraft detection/segmentation \cite{Pugliatti2021}, EfficientNet to detect satellites in the SPARK dataset \cite{aldahoul_localization_2022}, Faster R-CNN \cite{Sharma2020}, SSD \cite{Kaki2023, Lotti2023}, YOLOv3 \cite{Park2020}, YOLOv5 \cite{Piazza2022, Kaidanovic2022} to detect the RSO in the SPEED datasets.

While these results are sufficient for some applications, estimating the pose and locating entire spacecraft falls short of enabling autonomous docking with a non-cooperative spacecraft, as there remain collision risks with fragile components of a target. Hence, our reference mission \cite{mahendrakar2023autonomous} requires a finer-grained characterization of spacecraft components that can detect fragile components and identify safe docking points. Multiple works have pursued the satellite component detection problem--typically focusing on a subset of antennas, satellite bodies, solar panels, radiators, and thrusters. Several works used R-CNN \cite{Chen2020} and Faster R-CNN \cite{Viggh2023} to detect components of known satellites by training on synthetic \cite{Faraco2022} and real-life images \cite{Mahendrakar2022}. Satellite component detection for RPO applications must work in real-time using onboard computers to avoid lag times associated with ground control. Our prior work demonstrated Faster R-CNN is too computationally expensive for on-board use \cite{Mahendrakar2022}, and later works moved to more efficient single-stage object detectors, primarily YOLO-based methods \cite{Mahendrakar2024, Mahendrakar2021, Dung2021, Mahendrakar2021use, Mahendrakar2023}.


While these techniques highlight the power of CNNs in generalizability and their ability to learn patterns and similarities, they also acknowledge drawbacks such as misclassifications and missed detections, especially in scenarios with poor coverage in satellite image training datasets. However, many of the errors made by CNNs are easy for human experts to avoid on inspection. SpY leverages the strengths of CNNs while adding contextual knowledge through traditional computer vision techniques to encode human-like decision processes into satellite component detection. 



\section{DATASETS AND METHODS}

There are three distinct datasets used in this work. Web satellite dataset (WSD), shape detector dataset (SDD) and hardware in the loop (HIL) dataset. WSD and SDD are used for training and validation only while HIL is used for testing only.

\subsection{Web Satellite Dataset (WSD)}

\begin{figure}[ht]
    \centering
    \includegraphics[width=\columnwidth]{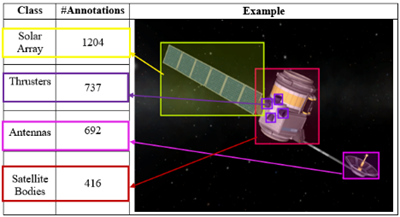}
    \caption{WSD Image Dataset}
    \label{fig:WSD}
\end{figure}

WSD, as described in \cite{Mahendrakar2024}, consists of both real and synthetic images of spacecraft sourced from the Internet. The selection criteria for these images are as follows:
\begin{enumerate}
    \item The objects must be identifiable, with each component distinguishable from the others 
    \item The shape of each component must accurately represent a real spacecraft component.
    \item Images in the dataset must not be repeated.
\end{enumerate}
The WSD contains a total of 1,231 images, each labeled for antennas, body panels, solar panels, and thrusters. All components also have bounding box annotations as illustrated in Fig.~\ref{fig:WSD}. The dataset is split into 80\% training and 20\% testing images.

\subsection{Shape Detector Dataset (SDD)}

The shape detector concept was first introduced in SpaceYOLO \cite{SpaceYOLO} to train YOLOv5 to identify primitive shapes such as circles and rectangles. However, the shape dataset used in SpaceYOLO lacked triangles and rings, other commonly occurring shapes in the SpaceYOLO survey. Furthermore, the original dataset was built manually. This work introduces an automated shape generator tool using the open-source Pycairo 2D graphics library that generates images with 2D circles, rectangles, triangles, and rings complete with bounding box and shape class annotations.

The shapes are printed individually in frames as well as printed together as collages. The shape generator outputs 2D shapes in 640px-by-640px frames and annotation boxes attached to each shape. It randomly selects gray, white, or black backgrounds and assigns the shapes different hues of gray. The SDD includes images with circles and rings with radii 5-10\% of the frame size, rectangles with widths and heights 5-50\%, and triangles with side lengths 5-10\%.

\begin{figure}[ht]
    \centering
    \includegraphics[width=\columnwidth]{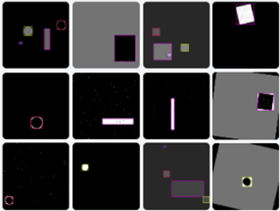}
    \caption{SDD}
    \label{fig:SDD_Dataset}
\end{figure}

Once the images and labels are generated, they are augmented with random rotations, shears, blurs, and noise. A sample of this dataset is shown in Fig.~\ref{fig:SDD_Dataset}.

\subsection{HIL Dataset}

\begin{figure}[ht]
    \centering
    \begin{subfigure}[b]{0.47\columnwidth}
        \centering
        \includegraphics[width=\columnwidth]{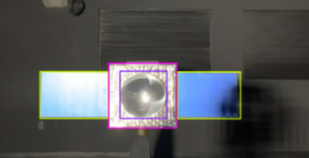} 
        \caption{Horizontal Solar Panels (HSP)}
        \label{fig:HSP}
    \end{subfigure}
    \hfill
    \begin{subfigure}[b]{0.47\columnwidth}
        \centering
        \includegraphics[width=\columnwidth]{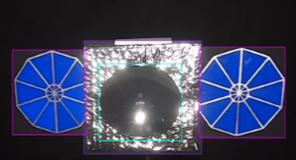} 
        \caption{Decagonal Solar Panels (DSP)}
        \label{fig:DSP}
    \end{subfigure}
    \vspace{.25cm}
    \begin{subfigure}[b]{0.48\columnwidth}
        \centering
        \includegraphics[width=\columnwidth]{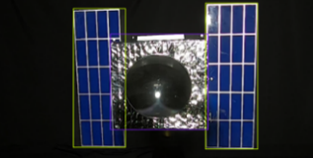} 
        \caption{Longitudinal Solar Panels (LSP)}
        \label{fig:LSP}
    \end{subfigure}
    \caption{HIL Dataset \cite{Mahendrakar2023}}
    \label{fig:HIL}
\end{figure}

The ORION testbed \cite{Wilde2016} at the Autonomy Lab in Florida Tech was used to generate HIL images \cite{Mahendrakar2023}. The testbed features a maneuver kinematics platform hosting two vehicles, with one on a gantry capable of moving in x and y directions. Both vehicles can pitch +/-30° and yaw infinitely, with one serving as the target satellite (mock-up) referred to as the resident space object (RSO). The RSO has configurable solar panels, antennas, and thrusters, which can be easily swapped out. The satellite body is wrapped in a material that looks like commonly-used multi-layer insulation (MLI). For this work, the solar panels were interchanged among decagonal, horizontal, and longitudinal configurations while leaving the rest of the features unchanged as shown in Fig.~\ref{fig:HIL}.

The lab environment itself features highly absorbent black paint on windows, doors, ceiling, and floors. Artificial sunlight is created using a Hilio D12 LED litepanel with adjustable power from 0\% to 100\%, generating a maximum intensity of 5600K daylight balanced temperature.

\begin{figure}[ht]
    \centering
    \begin{subfigure}[b]{0.47\columnwidth}
        \centering
        \includegraphics[height=3cm]{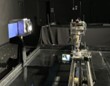} 
        \caption{Test Bed}
        \label{fig:testbed}
    \end{subfigure}
    \hfill
    \begin{subfigure}[b]{0.47\columnwidth}
        \centering
        \includegraphics[height=3cm]{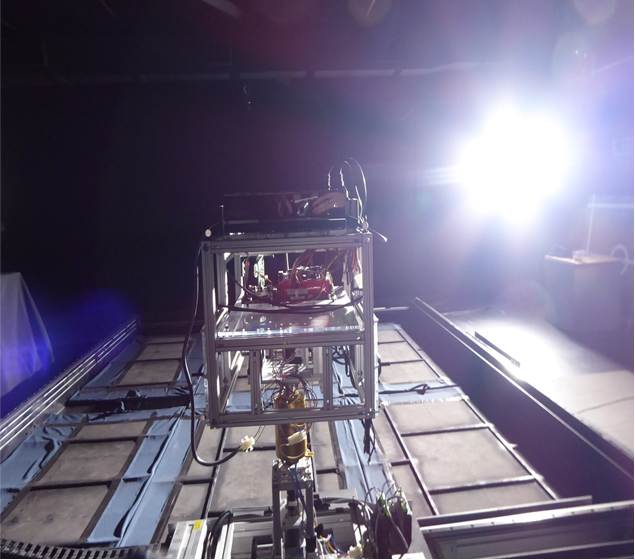} 
        \caption{Lighting Conditions}
        \label{fig:lighting}
    \end{subfigure}
    \caption{ORION Testbed \cite{Wilde2016}}
    \label{fig:ORION}
\end{figure}

Using the dynamic lighting capability, each mock-up configuration was subjected to four different lighting conditions, and videos of the RSO rotating and the chaser rendezvous approach are summarized in Table~\ref{tab:simulation_parameters}. Images from the videos are extracted at 1 frame per second and all visible solar panels, body panels, antenna and thruster are annotated.

\begin{table}[ht]
    \centering
     \caption{Simulation Parameters and Observer Distance}
    \begin{tabular}{ccccccc}
        \toprule
        Sim \# & Lighting & Light & Casting & Yaw & \multicolumn{2}{c}{Observer} \\
        & Intensity & Height & Angle & Rate & \multicolumn{2}{c}{Distance}\\
        \cmidrule(lr){6-7}
         & & & & [°/s] & X [m] & Y [m] \\
        \midrule
        1 & 20\% & 65" & 0° & 5 & 3.5 & 0 \\
        2 & 60\% & 40" & 30° & 5 & 3.5 & 0 \\
        3 & 5\% & 40" & 45° & 5 & 3.5 & 0 \\
        4 & 5\% & 40" & 0° & 5 & 3.5 & 0 \\
        \bottomrule
    \end{tabular}
    \label{tab:simulation_parameters}
\end{table}

\subsection{Methods}

This work compares several variations of SpY and SYC (described in full details in Section V) with baseline models from the literature, each trained on WSD and/or SDD. Further, an ensemble that combines SpY with a standard CNN-based object detector will be evaluated.

Each method includes an object detector and may or may not include a standalone classifier that contributes to class predictions. Object detectors include YOLO (YOLOv5 \cite{Jocher2020} trained on the WSD) and the shape detector (YOLOv5 trained on the SDD). YOLOv5 is selected since it demonstrates the best performance among comparable algorithms on HIL with sufficient framerates on current spaceflight-like hardware \cite{Mahendrakar2024}. Secondary classifiers include MobileNetV2 \cite{sandler2018mobilenetv2} and SYC. MobileNetV2 is selected because it is a lightweight architecture designed for low-SWaP hardware.

The methods evaluated in this work are described in Table~\ref{tab:methods}. All are compared on their performance in detecting components in HIL images not seen during training. 

\begin{table}[ht]
    \centering
    \caption{Methods Evaluated}
    \begin{tabular}{lcc}
        \toprule
        Model & Detector & Classifier \\
        \midrule
        YOLO & YOLOv5 on WSD & N/A \\
        YOLO+MN & YOLOv5 on WSD & MobileNetV2 \\
        YOLO+SYC \textbf{(ours)} & YOLOv5 on WSD & SYC \\
        SpY \textbf{(ours)} & Shape Detector & SYC \\
        SpY+YOLO \textbf{(ours)} & Ensemble & Ensemble\\
        \bottomrule
    \end{tabular}
    \label{tab:methods}
\end{table}

\section{SPACEYOLOV2 (SpY) OVERVIEW}


This section provides a detailed overview of the SpY architecture. Like any object detection architecture, SpY takes an image as an input and outputs predicted bounding boxes that localize and classify objects present in the image. Specifically, SpY identifies antennas, bodies, solar panels, thrusters, and unknown objects. It is further equipped to identify white radiators or other user-defined components, but this functionality is not measured in this work due to a lack of real-world testing data.

The unknown object class enables SpY to conclude a well-defined feature exists in a predicted region without making a class prediction. This ensures a component that cannot be definitively classified will not be misclassified. In the context of downstream navigation and guidance tasks, this serves as a safety feature.

\begin{figure*}[ht]
    \centering
    \includegraphics[width=\textwidth]{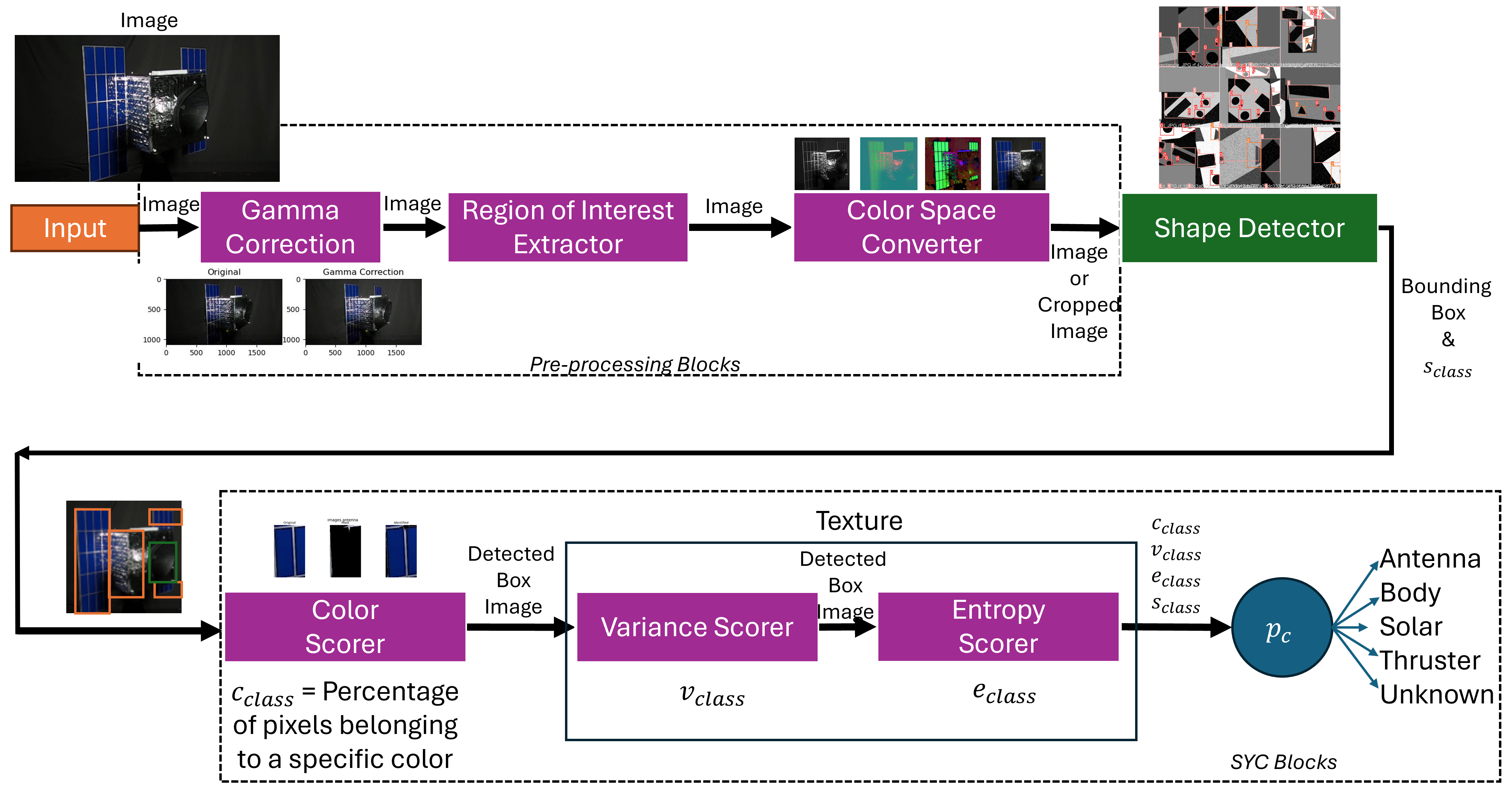} 
    \caption{SpaceYOLOv2 (SpY) Architecture}
    \label{fig:spy}
\end{figure*}

Shown in Fig.~\ref{fig:spy}, The SpY architecture begins with pre-processing blocks, followed by the shape detector that identifies and localizes shapes in the images. Next is SYC, which first uses specialized feature extractors to compute shape, color, and texture class scores for each shape’s bounding box in the original image. SYC then encodes human-like reasoning based on these features to classify the shapes as specific spacecraft components. 

Each of these three main parts of SpY are discussed in the forthcoming subsections.

\subsection{Pre-processing Blocks}

The pre-processing steps include gamma correction, region of interest (ROI) extraction, and color space conversion. Each block can be turned on or off as needed and the color space converter block supports the four color spaces (HSV, RGB, YCbCr, grayscale) as needed for individual applications.

\subsubsection{Gamma Correction}

Gamma correction \cite{Gonzalez2009} is a nonlinear operation used to encode and decode luminance in an image, enhancing contrast. This is especially useful for spacecraft in low lighting conditions to better define the edges of geometry. For spacecraft imagery sensitive to lighting, a threshold of $\gamma=0.8$ was selected to brighten the images. However, the gamma value does not dynamically change with the sun’s reflection angle on the spacecraft, which is a limitation that future work will address.

\subsubsection{ROI Extractor}
\begin{figure}
    \centering
    \includegraphics[width=\columnwidth]{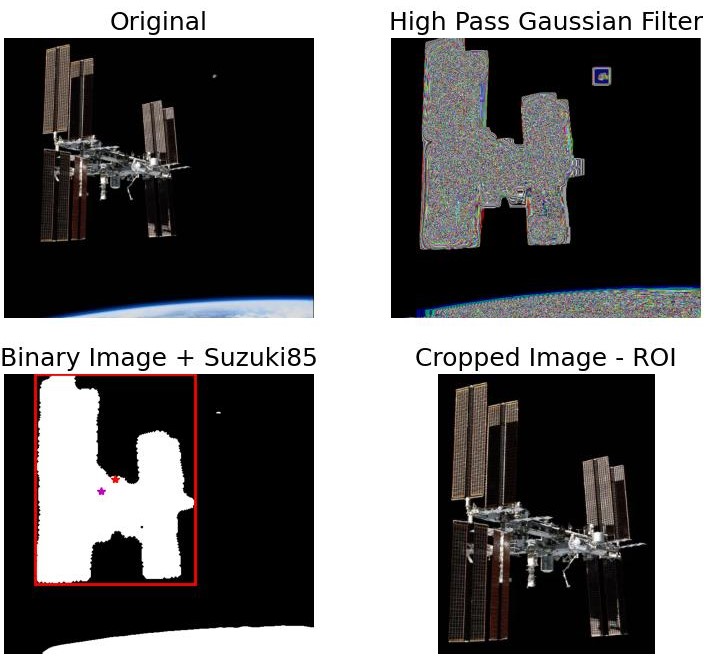}
    \caption{ROI Extractor}
    \label{fig:ROI_extractor}
\end{figure}

The image frame could have background details like the Earth or another spacecraft. In the HIL dataset, there is background clutter in the lab that can affect object detector performance. Our approach uses a high-pass Gaussian filter for background subtraction and the Suzuki85 \cite{Suzuki1985} contour detection algorithm to segment out the RSO, and extract a ROI tightly focused on the RSO. This is shown in Fig.~\ref{fig:ROI_extractor}.

To avoid clipping out important details, we ensured the ROI extractor does not eliminate any area contained in the ground truth bounding boxes from any image in the HIL datasets.

\subsubsection{Color Space Converter}

If the input image has 3 channels, the color space converter can convert the image or the cropped image (if the ROI extractor is on) into one of four user-selected color spaces: grayscale, YCbCr, HSV, or RGB. The output from this block is directly fed into the shape detector.

Each of these color spaces has unique properties and advantages in terms of separating chroma content (grayscale, YCbCr), luminance (YCbCr), or decoupling hue, saturation, and value (HSV). Depending on the imagery for the individual application of SpY, the choice of color space for images fed to the shape detection could significantly impact object detection performance.



\subsection{Shape Detector}


Typical YOLO models \cite{Redmon2016} optimize a loss function to learn how to predict the boxes using three parts. The bounding box loss ($L_\text{bbox}$) encourages accurate component localization. The objectness loss ($L_\text{obj}$) ensures the model predicts bounding boxes that contain objects. The classification performance ($L_\text{cls}$) encourages correct class predictions. The loss is a weighted sum of these with hyperparameters $\lambda_1$ and $\lambda_2$:
\begin{align}
    L=\lambda_1 L_\text{bbox} + \lambda_2 L_\text{obj} + L_\text{cls}
\end{align}

SpY takes a different approach. Its shape detector is trained on 2D shape images from the SDD and tested on spacecraft images. Unlike ordinary YOLO, the goal of the shape detector is not to directly detect spacecraft components but rather to detect shape primitives within the region bounded by the spacecraft's silhouette. These detected bounding boxes will be classified by their satellite component class by the context-based SYC discussed in full details in the next section.

Our goals extend beyond high-quality detection and is further concerned with detecting all components that are present. Therefore, we modify the YOLO loss to train the shape detector to predict bounding boxes covering of all components. Since our downstream classifier can label an object as ``unknown,'' this effectively reduces missed detections without increasing false positives.  This conservative approach ensures SpY's predictions are safe for use in downstream visual navigation tasks.

For each image processed by the the shape detector, the percentage of ground truth boxes detected as shapes. This performance ratio is termed as the shape detector overlap denoted by $SD_\text{overlap}$, computed for each image as:
\begin{align}
    SD_\text{overlap}=\frac{GT\cap\bigcup\limits_{i=1}^{n_{\text{pred}}}p_i}{GT}
\end{align}
where $n_\text{pred}$ is the number of of predicted boxes, $p_i$ are the predicted boxes, and GT is the union of all ground truth boxes.

\begin{figure}[ht]
    \centering
    \begin{subfigure}[b]{0.9\columnwidth}
        \centering
        \includegraphics[width=\columnwidth]{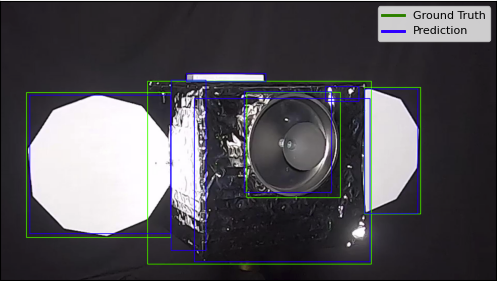} 
        \caption{Ground Truth and Predicted Boxes}
        \label{fig:gt_pred_boxes}
    \end{subfigure}
    \vspace{.25cm}
    \begin{subfigure}[b]{0.47\columnwidth}
        \centering
        \includegraphics[height=2.2cm]{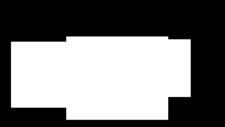} 
        \caption{Ground Truth Mask}
        \label{fig:gtmask}
    \end{subfigure}
    \hfill
    \begin{subfigure}[b]{0.47\columnwidth}
        \centering
        \includegraphics[height=2.2cm]{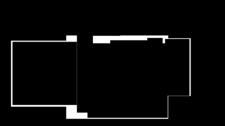} 
        \caption{Overlapping Area}
        \label{fig:detmask}
    \end{subfigure}
    \caption{SD Overlap}
    \label{fig:SD_overlap}
\end{figure}

Fig.~\ref{fig:SD_overlap} visualizes $SD_\text{overlap}$. GT is shown as white mask in Fig.~\ref{fig:gtmask} and portion of it covered bu detected bounding boxes is black in Fig.\ref{fig:detmask}). In this case, the shape detector's predictions overlap nearly all of the ground truth bounding boxes, indicating good coverage.

The standard data-driven loss function of YOLO is modified to subtract the mean $SD_\text{overlap}$ in each training batch to penalize the shape detector for failing to detect portions of the ground truth bounding boxes:
\begin{align}
    L_\text{SD}=L-\frac{\lambda_3}{m}\sum\limits_{i=1}^m SD_{\text{overlap},i}
\end{align}
where $m$ is the training batch size. We use $\lambda_3=1$.

\subsection{SpaceYOLOv2 Classifier (SYC)}

Inspired by a survey of aerospace professionals \cite{SpaceYOLO}, we next encode human reasoning into the pipeline by computing class scores to the bounding boxes detected by the shape detector based on shape, color, and texture. These scores are used by SYC to assign the final class predictions to the bounding boxes from the shape detector. 

\subsubsection{Shape Scorer}

Shapes are important for human reasoning about satellite components. Solar panels are typically rectangular and thin, corresponding to rectangular and ring shapes. Antennas are circular and concave, matching ring and circular geometries. The body is a cuboid or cylindrical structure, indicating a rectangular shape, while thrusters are conical (triangular) and ring-shaped, corresponding to triangles and rings.

The shape scorer incorporates contextual shape-based knowledge by assigning a shape class score $s_\text{class}$ to bounding box predictions from the shape detector. The scores use 1s to indicate that a shape can be any component, while 2s emphasize the most likely components. For example, $s_\text{thruster} = 2$ for a detected triangle since it is most likely a thruster. Full details are shown in Table~\ref{tab:shape_scores}.

\begin{table}[ht]
    \centering
    \begin{tabular}{lccccc}
        \toprule
        & Antenna & Body & Solar & Thruster & Unknown \\
        \midrule
        Circle & 2 & 1 & 1 & 1 & 1 \\
        Rectangle & 1 & 2 & 2 & 1 & 1 \\
        Triangle & 1 & 1 & 1 & 2 & 1 \\
        Ring & 2 & 1 & 2 & 2 & 1 \\
        \bottomrule
    \end{tabular}
    \caption{Shape Class Scores $s_\text{class}$}
    \label{tab:shape_scores}
\end{table}

\subsubsection{Color Scorer}

Another key cue for human satellite component detection is color. For example, blue objects are likely solar arrays and silver objects are more likely to be bodies or antennas. The color scorer extracts predicted bounding boxes from the original image and analyzes their color information to encode this simple reasoning by assigning color class scores $c_\text{class}$ to each bounding box.

The bounding box is first converted to HSV color space since its decoupled hue, saturation and value make colors easy to distinguish. We define six colors based on HSV ranges that coincide with human perception: blue (for solar panels), white (radiators), silver (body), 2 different intensities of gray (gray1 for antenna/body and gray2 for thruster/body) and black (for background or unknown). These ranges were extracted from HIL images not used during training nor testing.

Bounding box pixels are then segmented into these six color ranges and we compute the percentage of each color $p_\text{color}$. These percentages are used to compute color class scores $c_\text{class}$ for each bounding box. The class scores are computed as mean percentage of colors associated with each feature as shown in Table~\ref{tab:color_scores}.

\begin{table}[ht]
    \centering
    \begin{tabular}{lll}
        \toprule
         Class & Colors & Formula for $c_\text{class}$ \\
         \midrule
         Antenna & silver, gray1 & $\frac{1}{2}(p_\text{silver} + p_\text{gray1})$ \\
         Body & silver, gray1, gray2 & $\frac{1}{3}(p_\text{silver} + p_\text{gray1} + p_\text{gray2})$ \\
         Solar array & blue & $p_\text{blue}$ \\
         Thruster & silver, gray2 & $\frac{1}{2}(p_\text{silver} + p_\text{gray1})$ \\ 
         White radiator & white & $p_\text{white}$ \\
         Unknown & black & $p_\text{black}$\\
        \bottomrule
    \end{tabular}
    \caption{Color Class Scores $c_\text{class}$}
    \label{tab:color_scores}
\end{table}

Since there is no white radiator in the HIL dataset and the back of the solar panels is white, the color scorer is modified to combine the white radiator probability with the solar panel probability if the white percentage is greater than 0.5 ($p_\text{white}>0.5)$ for our testing below. This modification ensures that the absence of white radiators in the dataset does not negatively impact the classification performance for the solar component.

\subsubsection{Texture Scorer}

The third feature commonly used by humans to detect satellite features is texture. The texture scorer extracts bounding boxes in labeled HIL images and converts them to grayscale. It then computes measures of texture common in image processing, variance and entropy of the pixel intensities \cite{Gonzalez2009}:
\begin{align}
    \sigma^2_\text{pixels}&=\frac{1}{n}\sum\limits_{i=1}^{n_\text{box}} \left(x_i-\bar{x}\right)^2\\
    h_\text{bbox}&=-1000\sum\limits_{i=1}^{n_\text{pixels}} x_i \log_2(x_i)
\end{align}
where $n_\text{pixels}$ is the number of pixels in the bounding box, $x_i$ is the pixel intensity, $\bar{x}$ is the mean pixel intensity. The entropy values are multiplied by 1000 to match the order of magnitude of the variance, ensuring that we maintain higher fidelity and avoid losing information due to truncation.


Both measures correspond to texture, but there are subtle differences \cite{Lhermitte2022, Gonzalez2009}. Variance indicates the degree of variation or contrast within the bounding box, which is effective for measuring homogeneous textures like smooth pixel intensity gradients with high-magnitude changes. Entropy is an effective measure of the degree to which there are high-frequency changes in pixel intensity, such as sharp boundaries and heterogeneous surfaces.

Next, we compute texture class scores for variance $v_\text{class}$ and entropy $e_\text{class}$. To establish the link between texture and object class, variance and entropy of annotated bounding boxes are computed and histograms for each class are developed based on real-world HIL images because they exhibit realistic pixel-level details, unlike some of the often over-smoothed synthetic images in WSD.

\begin{figure}[ht]
    \centering
    \begin{subfigure}[b]{0.49\columnwidth}
        \centering
        \includegraphics[width=\columnwidth]{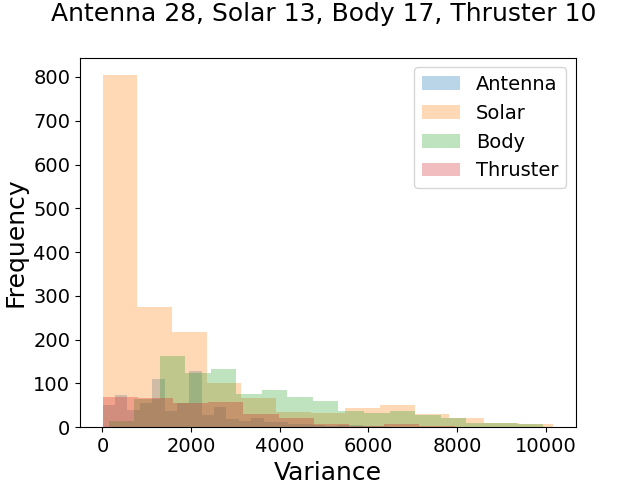} 
        \caption{Variance Class Histograms}
    \end{subfigure}
    \hfill
    \begin{subfigure}[b]{0.49\columnwidth}
        \centering
        \includegraphics[width=\columnwidth]{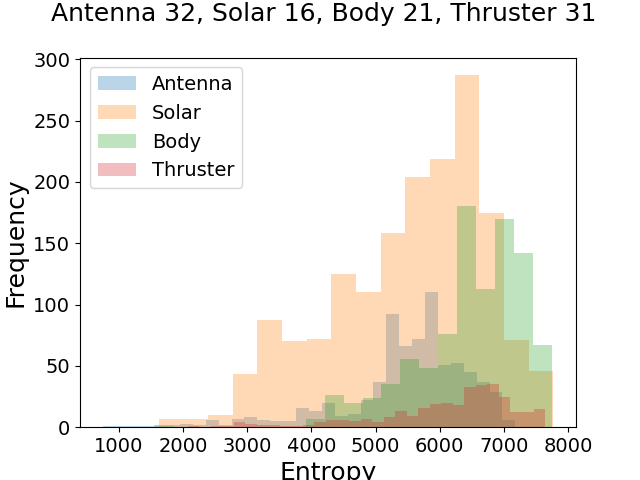} 
        \caption{Entropy Class Histograms}
    \end{subfigure}
    \caption{Texture Class Histograms}
    \label{fig:texture_histograms}
\end{figure}

Histograms are shown in Figure~\ref{fig:texture_histograms}. We note variance and entropy skew inversely to one another and class histograms exhibit different patterns, underscoring the complementary nature of the two measures.

After a performance comparison with bin sizes ranging from 10 to 100, it was determined the Freedman-Diconis rule \cite{Freedman1981} is optimal for our purposes. Hence, the number of uniform bins in each class histogram for each texture measure is
\begin{align}
    \text{Bin Count}=\frac{2IQR}{\sqrt[3]{n_\text{class}}}
\end{align}
where $n_\text{class}$ is the number of components of the class in the dataset and IQR is the interquartile range of the specified metric for the specified class. The numbers of bins span 0 to 10000 for variance and and 0 to 8000 for entropy.

For an input bounding box, variance $\sigma^2_\text{bbox}$ is computed and the corresponding bin is determined for each class histogram and the variance relative frequency $vr_\text{class}$ of the classes for that variance is computed as
\begin{align}
    vr_\text{class}(\sigma^2_\text{bbox})=\frac{f_\text{class}(\sigma^2_\text{bbox})}{\sum\limits_{i=1}^4 f_{\text{class}_i}(\sigma^2_\text{bbox})}
\end{align}
where classes assessed for texture consist of antenna, body, solar, and thruster. Entropy relative frequencies $er_\text{class}$ are computed similarly. In practice, we reduce compute costs by creating a look-up table for variances $\sigma^2_\text{bbox}\in[0,10000]$ and entropies $h_\text{bbox}\in[0,8000]$.

The HIL dataset used to develop these scores has an imbalanced number of components. There are 741 antennas, 1692 solar panels, 966 body annotations and 320 thrusters. Hence, solar panels dominate over other components. To remove this bias, texture class scores are multiplied by the ratio of the total number of solar panels (1692) to the number of objects present in the respective class to compute the final texture class scores:
\begin{align}
    v_\text{class}&=\frac{1692vr_\text{class}}{|\# \text{ objects in class}|}
    \\e_\text{class}&=\frac{1692er_\text{class}}{|\# \text{ objects in class}|}
\end{align}

\subsubsection{SYC Predictions}

At this stage, the shape detector has predicted bounding boxes classified by shape and the feature extractors have computed shape ($s_\text{class}$), color ($c_\text{class}$), and texture ($v_\text{class}$ and $e_\text{class}$) class scores for each bounding box. The last piece of SYC uses these contextual scores to predict the satellite component class for each bounding box through a rule-based approach.

Two voting techniques are used to combine the scores into a final set of class scores. They are predictive soft voting (PSV) and multi-voting (MUV):
\begin{align}
    \text{PSV}_\text{class}&=s_\text{shape}\left(c_\text{class}+v_\text{class}+e_\text{class}\right)\\
    \text{MUV}_\text{class}&=s_\text{shape}\left(v_\text{class}+e_\text{class}\right)c_\text{class}
\end{align}

SYC uses Algorithm~\ref{alg:SYC} to ensemble these class scores into a final class prediction for SpY.

\begin{algorithm}
\caption{SYC Ensemble}\label{alg:SYC}
\begin{algorithmic}[1]
\State \textbf{Input:} $s$ (shape scores), $c$ (color scores), $v$ (variance scores), $e$ (entropy scores)
\State \textbf{Output:} $\hat{c}$ (predicted class for bbox)\\

\State $PSV \gets  \texttt{compute\_psv}(s,c,v,e)$ 
\State $MUV \gets \texttt{compute\_muv}(s,c,v,e)$\\


\State $p \gets max\_PSV\_class$ \Comment{PSV class name}
\State $m \gets max\_MUV\_class$ \Comment{MUV class name}\\

\State $pp\gets \texttt{max}(PSV)$
\State $mp\gets \texttt{max}(MUV)$\\

\If{$p \text{ is unknown} \textbf{ and } pp>0.5$}
    \State $\hat{c}=\text{unknown}$
\ElsIf{$m \text{ is unknown} \textbf{ and } mp>0.5$}
    \State $\hat{c}=\text{unknown}$
\ElsIf{$max\_color\_score \text{ is blue}$}
    \State $\hat{c}=\text{solar}$
\ElsIf{$p \text{ is thruster}$}
    \State $\hat{c}=\text{thruster}$
\ElsIf{$p \text{ is antenna} \textbf{ and } pp>mp$}
    \State $\hat{c}=\text{antenna}$
\ElsIf{$max\_variance\_class \text{ is solar}$}
    \State $\hat{c}=\text{solar}$
\Else
    \State $\hat{c}=m$
\EndIf
\end{algorithmic}
\end{algorithm}

Alternative approaches are used for classification when YOLO is paired with SpY or SYC because YOLO tends to be significantly better at detecting satellite bodies. This is because the body often includes attached antennas, thrusters, or solar panels that, which obscure the shape of the body, or split it into what looks like several shapes. YOLO is not reliant on shape, color, or texture and is more able to reason from less interpretable cues present in the input images.

When SYC is used as a secondary classifier for YOLO in the YOLO+SYC method, SYC body class predictions are simply ignored.

Inspired by SatSplatYOLO \cite{Nguyen2024}, the SpY+YOLO ensemble combines the data-driven YOLO predictions and context-based SpY predictions as follows.
\begin{enumerate}
    \item Use YOLO detections for the body component and ignore SpY body predictions.
    \item For overlapping YOLO and SpY boxes (IoU $>$ 0.5), perform a confidence score-weighted average of the box centers and dimensions.
    \item Calculate the new confidence score as the mean of SpY and YOLO confidences.
    \item Retain other YOLO or SpY boxes as they are.
\end{enumerate}

This fuses the strengths of each algorithm by allowing YOLO to detect bodies and taking input from SpY/SYC only when it tends to outperform YOLO.

\section{RESULTS AND ANALYSIS}

This section discusses the metrics used to evaluate the methods and model training processes. Next is SpY hyperparameter tuning and experimental results. Last is an analysis of the strengths and failure modes of several variations of SpY as they relate to baseline methods.

\subsection{Metrics}


We use several metrics to evaluate the performance of SpY and the other methods in Table~\ref{tab:methods}. A true positive (TP) is defined as a predicted detection with sufficiently high intersection over union (IoU) with a ground truth bounding box that is classified correctly. Any other detection is a false positive (FP) and a false negative (FN) is a failure to detect a ground truth object. The counts of these types of detections allow us to compute metrics precision (P), recall (R), and F1 score:

\begin{align}
    P=\frac{TP}{TP+FP},\,\, R=\frac{TP}{TP+FN},\,\, F1=\frac{2*P*R}{P+R}
\end{align}

Precision is the fraction of positive detections that are correct. Recall is the fraction of ground truth objects that are correctly detected. F1 score is the geometric mean of precision and recall.

We additionally use the standard mAP@0.5 object detection metric as a one-number summary of overall object detection performance. mAP is the mean of the average precision (AP) calculated for $N$ classes and $AP_i$ is as the area under the precision-recall curve for class $i$. The 0.5 represents the intersection over union (IoU) thresholds required for a true positive \cite{Padilla2021}.  
\begin{equation}
\text{mAP} = \frac{1}{N} \sum_{i=1}^{N} \text{AP}_i
\end{equation}

While we seek high precision and mAP, we focus most strongly on recall because it measures if we misclassify satellite components or fail to detect them entirely. For our use-cases in navigation and guidance for close-proximity operations, failing to detect hazards correctly are the most detrimental errors.



\subsection{Training}

YOLO--i.e. YOLOv5 (small) trained on the WSD--has validation mAP@0.5 0.587 for detecting antennas, bodies, solar panels and thrusters \cite{Mahendrakar2024}. The shape detector is trained on grayscale SDD images with intensities replicated in 3 color channels. This enables the shape detector to run inference on different color spaces. The resulting shape detector has validation mAP@0.5 0.947 for detecting circles, rectangles, rings, triangles, and rings.

Further analysis compared $SD_\text{overlap}$ for YOLO and SpY architectures against the HIL dataset. The analysis reveals that both share similar overlap performance, indicating they are comparable. However, SpY has far more detections both inside (x1.32) and outside (x2.69) the ground truth area (region of interest), indicating more noisy detections with SpY than YOLO weights. This is because YOLO is trained to identify only spacecraft components, while SpY's shape detector is designed to identify any primitive geometry, making it sensitive to all objects, including non-relevant ones like gantry rails and curtain creases in the HIL dataset. 



For the YOLO+MN method, MobileNetV2 was trained on cropped images of components from the WSD dataset. For training, the classifier attained an accuracy of 0.89 and an F1-score of 0.89. 


\subsection{SpY Optimal Hyperparameters}

Using the trained YOLO weights, a grid search was conducted on HIL datasets to optimize YOLO+SYC. A total of 592 combinations of hyperparameters were tested: 37 pairs of entropy and variance bin sizes, ROI extractor (on/off), gamma corrector (on/off), and 4 color spaces for YOLO (at inference time). Metrics were computed across the 12 HIL datasets and 4 satellite component classes. Optimal hyperparameters were chosen by selecting the model with the highest F1 score without abnormally low scores on individual HIL datasets or components. 

The resulting model showed YOLO performs best in the RGB color space, consistent with its training data in the WSD. The optimal configuration for SYC uses the Freedman-Diaconis rule \cite{Freedman1981} for variance and entropy bin counts with neither gamma correction nor ROI extraction used in preprocessing.

For SpY, a separate a grid search of the pre-processing and shape detector hyperparameters was performed with the SYC hyperparmeters selected above. Best performance on our data is achieved without gamma correction or the ROI extractor preprocessing and grayscale processing for the shape detector, matching the grayscale training data in the SDD.

Applications with imagery different from ours could benefit from different settings. The public SpY codebase enables task-specific optimizations.

\subsection{Experimental Results}

This section provides experimental results comparing several variations of the context-based SpY/SYC with purely data-driven baseline models outlined in Table~\ref{tab:methods}. All quantitative results are summarized in Table~\ref{tab:performance}.

\begin{table*}[ht]
    \centering
    \footnotesize 
    \caption{Performance Analysis}
    \begin{tabular}{lccccccccccc}
        \toprule
        & \multicolumn{3}{c}{Standard Metrics} & \multicolumn{4}{c}{Class-level Recall} & \multicolumn{3}{c}{Solar Panel Type Recall} \\
        \cmidrule(lr){2-4} \cmidrule(lr){5-8} \cmidrule(lr){9-11}
        Method & F1 & mAP & R & Antenna & Body & Solar & Thruster & DSP & HSP & LSP \\
        \midrule
        YOLO & \underline{\textbf{0.315}} & \underline{\textbf{0.361}} & 0.362 & 0.157 & 0.555 & 0.363 & 0.010 & 0.191 & \textbf{0.313} & \underline{\textbf{0.585}} \\
        YOLO+MN & 0.145 & 0.150 & 0.214 & 0.024 & 0.534 & 0.120 & 0.008 & 0.018 & 0.080 & 0.264 \\
        YOLO+SYC \textbf{(ours)} & 0.268 & 0.257 & \textbf{0.381} & 0.019 & \underline{\textbf{0.613}} & \underline{\textbf{0.413}} & 0.017 & \underline{\textbf{0.596}} & 0.264 & \textbf{0.381} \\
        SpY \textbf{(ours)} & 0.167 & 0.131 & 0.210 & \textbf{0.178} & 0.198 & 0.286 & \textbf{0.051} & \textbf{0.362} & \textbf{0.387} & 0.109 \\
        SpY+YOLO \textbf{(ours)} & \textbf{0.304} & \textbf{0.280} & \underline{\textbf{0.436}} & \textbf{0.215} & \textbf{0.577} & \textbf{0.491} & \underline{\textbf{0.056}} & \textbf{0.426} & \underline{\textbf{0.479}} & \textbf{0.568} \\
        \bottomrule
    \end{tabular}
    \caption*{Top performers are \underline{\textbf{bolded and underlined}}. The second best are only \textbf{bolded}.}
    \label{tab:performance}
\end{table*}

The first column group of Table~\ref{tab:performance} includes F1, mAP@0.5, and recall of the methods across all satellite component classes. The ensemble of SpY and YOLO has the highest recall by a significant margin, indicating the fewest objects not detected. On the other hand, YOLO has the highest F1 and mAP scores, indicating higher precision and localization performance. SpY+YOLO is second best in these metrics with just 0.011 lower F1.

To investigate the performance of SpY further, component-wise recall analysis is tabulated for each model in the second set of columns of Table~\ref{tab:performance}. SpY has the lowest solar panel and body detections while the ensemble of SpY and YOLO has the highest recall. The low recall performance for solar panels and the body is due to SpY’s extreme sensitivity to shapes, unlike YOLO. SpY identifies even the smallest features, such as an OptiTrack marker, and classifies uncertain components as unknown, validating its fault tolerance.

SpY detects smaller rectangles in LSP and triangles in DSP. These are correctly classified as solar panels, but their IoU with ground truth boxes is low, so these are considered false positives. In contrast, DSP's decagonal structures are correctly identified and classified, resulting in true positives. As shown in the last column group in Table~\ref{tab:performance}, LSP detections are 3.3 times worse than DSP. Future work should include merging detections boxes present in a small, bounded region of connected sub-boxes indicating those boxes belong to the same feature to improve recall.

\begin{figure}[h]
    \centering
    \begin{subfigure}[b]{0.9\columnwidth}
        \centering
        \includegraphics[width=0.7\columnwidth]{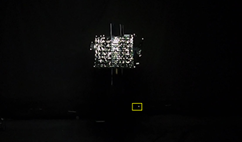} 
        \caption{OptiTrack Marker Detection}
        \label{fig:image1}
    \end{subfigure}
    \vspace{0.25cm}
    \begin{subfigure}[b]{0.47\columnwidth}
        \centering
        \includegraphics[height=2.2cm]{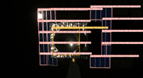} 
        \caption{LSP Detection}
        \label{fig:image2}
    \end{subfigure}
    \hfill
    \begin{subfigure}[b]{0.47\columnwidth}
        \centering
        \includegraphics[height=2.2cm]{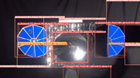} 
        \caption{DSP Detection}
        \label{fig:image3}
    \end{subfigure}
    \caption{SpY Shape Sensitivity}
    \label{fig:spy_shape_sesnsitivity}
\end{figure}

SpY has the worst recall performance for body due to its non-homogenous nature. Sometimes small rectangular regions between the edges of the body and solar panel get detected and classified correctly as body, but they are technically false positives, just like the small LSP solar panel detections. However, depending on the RPO application, knowing that body panels exist along with knowing what regions within it are free of other components is more valuable than detecting what may just be the central portion of the spacecraft. Such ``false positives'' would be beneficial in this context.

\subsection{Misclassification Analysis}

Next, we analyze incorrect classifications. Misclassifications are tallied in Table~\ref{tab:misclassification_analysis}. The first column shows total misclassifications for each method and the second column total misclassifications, excluding bodies.

\begin{table}[ht]
    \centering
    \caption{Misclassification Analysis}
    \begin{tabular}{lccc}
        \toprule
        Method & All Classes & No Body \\
        \midrule
        YOLO & \underline{\textbf{942}}& 694 \\
        YOLO+MN & 1666& 748 \\
        YOLO+SYC \textbf{(ours)} & 1560 & \textbf{333} \\
        SpY \textbf{(ours)} & 1038& \underline{\textbf{307}} \\
        SpY+YOLO \textbf{(ours)} & \textbf{977}& 594 \\
        \bottomrule
    \end{tabular}
    \caption*{Top performers are \underline{\textbf{bolded and underlined}}. The second best are only \textbf{bolded}.}
    \label{tab:misclassification_analysis}
\end{table}

When all classes are considered, YOLO and SpY+YOLO have the fewest misclassifications. However, when the body class is suppressed, SpY/SYC methods have far fewer misclassifications than the data-driven YOLO methods.

In summary, the experimental results and error analysis indicate SpY can detect components, but the combination of SpY with an additional CNN-based object detector such as the ensemble of YOLO and SpY is generally most effective.


Lastly, we discuss the use of secondary classifiers with YOLO. MobileNetV2 does not improve YOLO's performance by any metric. On the other hand, SYC boosts YOLO's recall and precision (excluding body) in numerous cases with its context-based analysis. Since YOLO simultaneously learns to predict bounding boxes and object classes, it reasons globally from entire input images. In contrast, MobileNetV2 reasons solely from pixels within bounding boxes extracted by YOLO, losing the capacity to reason globally. We hypothesize such global reasoning is critical to CNN classification performance but less so for SYC, which only needs local information.

\section{CONCLUSION}

This work presents an end-to-end context-based satellite component detector, SpY, that infuses modern CNN-based methods with human-inspired reasoning capabilities for increased accuracy and fault tolerance features for downstream visual navigation and guidance tasks. SpY is made up of a YOLOv5 object detector trained to detect primitive shapes within input imagery and SYC that leverages color and texture information to classify those shapes as antennas, satellite bodies, solar arrays, thrusters, or ``unknown.''

SpY demonstrated it can effectively identify spacecraft components while reasoning its detections. For example, it detects the antenna by first identifying a circular object and further classifying it based on shape, color, and texture features.

Test performance comparisons of various models on HIL data revealed that SpY is very sensitive to shapes, leading to an increased number of false positives. However, an ensemble SpY with YOLO has similar object detection performance with significantly higher recall than YOLO itself. Further SpY's capacity to label detections as unknown allows it to avoid feeding incorrect information to guidance and navigation systems. These advantages establish the YOLO+SpY ensemble is significantly more fault tolerant than purely data-driven methods.

Further, while CNN-based object detectors are very effective, they are not easily explainable. Methods like PEEK \cite{Meni2024} and Grad-CAM \cite{Selvaraju2020} have made strides in understanding how CNNs make their decisions by finding patterns in hidden states of neural networks with reference to the input pixel regions. In fact PEEK is class agnostic unlike GradCAM and has been used to look into the CNN layers of the YOLO model.  The step-by-step SpY decisions provide interpretability that refers to simple features like shape, color, and texture. Ongoing research in hybrid CNN/rules-based vision systems like SpY should cross-reference these complementary approaches (PEEK in case of SpY due to its class agnostic nature) to enhance the explainability and design of in-space computer vision systems. PEEK could also help prune the shape detector's CNN layers significantly to make it an even more computationally efficient for SWaP hardware than it already is. 


The SpY+YOLO ensemble has the capacity to combine human-guided contextual detection and pure CNN-based detections. It is the best prospect for safe and autonomous vision-based navigation for RPO around unknown satellites. 

\section*{ACKNOWLEDGMENT}

The authors would like to thank Drew Takeda, Markus Wilde, Mackenzie Meni, Minh Nguyen, Andrew Ekblad, Steven Wyatt, Nehru Attzs and Seema Putane for their helpful comments prior to submission.


\bibliographystyle{IEEEtran}
\bibliography{SpYNew}

\begin{thebibliography}{10}
\providecommand{\url}[1]{#1}
\csname url@samestyle\endcsname
\providecommand{\newblock}{\relax}
\providecommand{\bibinfo}[2]{#2}
\providecommand{\BIBentrySTDinterwordspacing}{\spaceskip=0pt\relax}
\providecommand{\BIBentryALTinterwordstretchfactor}{4}
\providecommand{\BIBentryALTinterwordspacing}{\spaceskip=\fontdimen2\font plus
\BIBentryALTinterwordstretchfactor\fontdimen3\font minus \fontdimen4\font\relax}
\providecommand{\BIBforeignlanguage}[2]{{%
\expandafter\ifx\csname l@#1\endcsname\relax
\typeout{** WARNING: IEEEtran.bst: No hyphenation pattern has been}%
\typeout{** loaded for the language `#1'. Using the pattern for}%
\typeout{** the default language instead.}%
\else
\language=\csname l@#1\endcsname
\fi
#2}}
\providecommand{\BIBdecl}{\relax}
\BIBdecl

\bibitem{Mahendrakar2024}
T.~Mahendrakar, R.~T. White, M.~Tiwari, and M.~Wilde, ``Unknown non-cooperative spacecraft characterization with lightweight convolutional neural networks,'' \emph{Journal of Aerospace Information Systems}, vol.~21, 2024.

\bibitem{Mahendrakar2021}
T.~Mahendrakar, R.~T. White, M.~Wilde, B.~Kish, and I.~Silver, ``Real-time satellite component recognition with yolo-v5,'' \emph{Small Satellite Conference}, 2021.

\bibitem{Mahendrakar2022}
T.~Mahendrakar, A.~Ekblad, N.~Fischer, R.~White, M.~Wilde, B.~Kish, and I.~Silver, ``Performance study of yolov5 and faster r-cnn for autonomous navigation around non-cooperative targets,'' vol. 2022-March, 2022.

\bibitem{Cutler2022}
J.~Cutler, M.~Wilde, A.~Rivkin, B.~Kish, and I.~Silver, ``Artificial potential field guidance for capture of non-cooperative target objects by chaser swarms,'' vol. 2022-March, 2022.

\bibitem{mahendrakar2023autonomous}
T.~Mahendrakar, S.~Holmberg, A.~Ekblad, E.~Conti, R.~T. White, M.~Wilde, and I.~Silver, ``Autonomous rendezvous with non-cooperative target objects with swarm chasers and observers,'' 2023.

\bibitem{SpaceYOLO}
T.~Mahendrakar, R.~T. White, M.~Wilde, and M.~Tiwari, ``Spaceyolo: A human-inspired model for real-time, on-board spacecraft feature detection,'' in \emph{2023 IEEE Aerospace Conference}, 2023, pp. 01--11.

\bibitem{Goodman2006}
J.~L. Goodman, ``History of space shuttle rendezvous and proximity operations,'' \emph{Journal of Spacecraft and Rockets}, vol.~43, 2006.

\bibitem{Yoshida2004}
K.~Yoshida, ``Engineering test satellite vii flight experiments for space robot dynamics and control: Theories on laboratory test beds ten years ago, now in orbit,'' \emph{The International Journal of Robotics Research}, vol.~22, 2004.

\bibitem{Davis2004}
T.~M. Davis and D.~Melanson, ``Xss-10 microsatellite flight demonstration program results,'' vol. 5419, 2004.

\bibitem{afrl_xss-11_2011}
\BIBentryALTinterwordspacing
{AFRL}, ``{XSS}-11 {Micro} {Satellite},'' Tech. Rep., 2011. [Online]. Available: \url{https://www.kirtland.af.mil/Portals/52/documents/AFD-111103-035.pdf?ver=2016-06-28-110256-797}
\BIBentrySTDinterwordspacing

\bibitem{afrl_automated_2014}
\BIBentryALTinterwordspacing
AFRL, ``Automated {Navigation} and {Guidance} {Experiment} for {Local} {Space} ({ANGELS}),'' Tech. Rep., 2014. [Online]. Available: \url{https://www.kirtland.af.mil/Portals/52/documents/AFD-131204-039.pdf?ver=2016-06-28-105617-297}
\BIBentrySTDinterwordspacing

\bibitem{Kennedy2008}
F.~G. Kennedy, ``Orbital express: Accomplishments and lessons learned,'' vol. 131, 2008.

\bibitem{Pyrak2022}
M.~Pyrak and J.~Anderson, ``Performance of northrop grumman’s mission extension vehicle (mev) rpo imagers at geo,'' 2022.

\bibitem{Ekblad2023}
A.~Ekblad, T.~Mahendrakar, R.~White, M.~Wilde, I.~Silver, and B.~Wheeler, ``Resource-constrained fpga design for satellite component feature extraction,'' vol. 2023-March, 2023.

\bibitem{Black2021}
K.~Black, S.~Shankar, D.~Fonseka, J.~Deutsch, A.~Dhir, and M.~R. Akella, ``Real-time, flight-ready, non-cooperative spacecraft pose estimation using monocular imagery,'' \emph{AAS}, 1 2021.

\bibitem{Hogan2021}
M.~Hogan, D.~Rondao, N.~Aouf, and O.~Dubois-Matra, ``Using convolutional neural networks for relative pose estimation of a non-cooperative spacecraft with thermal infrared imagery,'' \emph{European Space Agency Guidance, Navigation and Control Conference}, 5 2021.

\bibitem{Rondao2023}
D.~Rondao, N.~Aouf, and M.~A. Richardson, ``Chinet: Deep recurrent convolutional learning for multimodal spacecraft pose estimation,'' \emph{IEEE Transactions on Aerospace and Electronic Systems}, vol.~59, 2023.

\bibitem{Garcia2021}
A.~Garcia, M.~A. Musallam, V.~Gaudilliere, E.~Ghorbel, K.~A. Ismaeil, M.~Perez, and D.~Aouada, ``Lspnet: A 2d localization-oriented spacecraft pose estimation neural network,'' 2021.

\bibitem{Zhou2022}
D.~Zhou, G.~Sun, W.~Lei, and L.~Wu, ``Space noncooperative object active tracking with deep reinforcement learning,'' \emph{IEEE Transactions on Aerospace and Electronic Systems}, vol.~58, 2022.

\bibitem{Speed}
\BIBentryALTinterwordspacing
S.~Sharma, T.~H. Park, and S.~D'Amico, ``Spacecraft pose estimation dataset (speed),'' \emph{Stanford Digital Repository}. [Online]. Available: \url{https://purl.stanford.edu/dz692fn7184.}
\BIBentrySTDinterwordspacing

\bibitem{Park2022}
T.~H. Park, M.~Martens, G.~Lecuyer, D.~Izzo, and S.~D'Amico, ``Speed+: Next-generation dataset for spacecraft pose estimation across domain gap,'' vol. 2022-March, 2022.

\bibitem{Aarestad2020}
J.~Aarestad, A.~Cochrane, M.~Hannon, E.~Kain, C.~Kief, S.~Lindsley, and B.~Zufelt, ``Challenges and opportunities for cubesat detection for space situational awareness using a cnn,'' \emph{Small Satellite Conference}, 2020.

\bibitem{Pugliatti2021}
M.~Pugliatti and F.~Topputo, ``Navigation about irregular bodies through segmentation maps,'' \emph{AAS 21-383}, 2021.

\bibitem{aldahoul_localization_2022}
\BIBentryALTinterwordspacing
N.~AlDahoul, H.~A. Karim, A.~De~Castro, and M.~J.~T. Tan, ``Localization and classification of space objects using {EfficientDet} detector for space situational awareness,'' \emph{Scientific Reports}, vol.~12, no.~1, p. 21896, Dec. 2022. [Online]. Available: \url{https://doi.org/10.1038/s41598-022-25859-y}
\BIBentrySTDinterwordspacing

\bibitem{Sharma2020}
S.~Sharma and S.~D'Amico, ``Neural network-based pose estimation for noncooperative spacecraft rendezvous,'' \emph{IEEE Transactions on Aerospace and Electronic Systems}, vol.~56, pp. 4638--4658, 12 2020.

\bibitem{Kaki2023}
S.~Kaki, J.~Deutsch, K.~Black, A.~Cura-Portillo, B.~A. Jones, and M.~R. Akella, ``Real-time image-based relative pose estimation and filtering for spacecraft applications,'' \emph{Journal of Aerospace Information Systems}, vol.~20, 2023.

\bibitem{Lotti2023}
A.~Lotti, D.~Modenini, P.~Tortora, M.~Saponara, and M.~A. Perino, ``Deep learning for real-time satellite pose estimation on tensor processing units,'' \emph{Journal of Spacecraft and Rockets}, vol.~60, 2023.

\bibitem{Park2020}
T.~H. Park, S.~Sharma, and S.~D’amico, ``Towards robust learning-based pose estimation of noncooperative spacecraft,'' vol. 171, 2020.

\bibitem{Piazza2022}
M.~Piazza, M.~Maestrini, and P.~D. Lizia, ``Monocular relative pose estimation pipeline for uncooperative resident space objects,'' \emph{Journal of Aerospace Information Systems}, vol.~19, 2022.

\bibitem{Kaidanovic2022}
D.~Kaidanovic, M.~Piazza, M.~Maestrini, and P.~D. Lizia, ``Deep learning-based relative navigation about uncooperative space objects,'' vol. 2022-September, 2022.

\bibitem{Chen2020}
Y.~Chen, J.~Gao, and K.~Zhang, ``R-cnn-based satellite components detection in optical images,'' \emph{International Journal of Aerospace Engineering}, vol. 2020, 2020.

\bibitem{Viggh2023}
H.~Viggh, S.~Loughran, Y.~Rachlin, R.~Allen, and J.~Ruprecht, ``Training deep learning spacecraft component detection algorithms using synthetic image data,'' vol. 2023-March, 2023.

\bibitem{Faraco2022}
N.~Faraco, M.~Maestrini, and P.~D. Lizia, ``Instance segmentation for feature recognition on noncooperative resident space objects,'' \emph{Journal of Spacecraft and Rockets}, vol.~59, 2022.

\bibitem{Dung2021}
H.~A. Dung, B.~Chen, and T.~J. Chin, ``A spacecraft dataset for detection, segmentation and parts recognition,'' 2021.

\bibitem{Mahendrakar2021use}
T.~Mahendrakar, J.~Cutler, N.~Fischer, A.~Rivkin, A.~Ekblad, K.~Watkins, M.~Wilde, R.~White, B.~Kish, and I.~Silver, ``Use of artificial intelligence for feature recognition and flightpath planning around non-cooperative resident space object,'' 2021.

\bibitem{Mahendrakar2023}
T.~Mahendrakar, M.~N. Attzs, A.~L. Tisaranni, J.~M. Duarte, R.~T. White, and M.~Wilde, ``Impact of intra-class variance on yolov5 model performance for autonomous navigation around non-cooperative targets.''\hskip 1em plus 0.5em minus 0.4em\relax AIAA SciTech, 2023.

\bibitem{Wilde2016}
M.~Wilde, B.~Kaplinger, T.~Go, H.~Gutierrez, and D.~Kirk, ``Orion: A simulation environment for spacecraft formation flight, capture, and orbital robotics,'' vol. 2016-June, 2016.

\bibitem{Jocher2020}
G.~Jocher, ``Yolo v5 by ultralytics,'' 2020.

\bibitem{sandler2018mobilenetv2}
M.~Sandler, A.~Howard, M.~Zhu, A.~Zhmoginov, and L.-C. Chen, ``Mobilenetv2: Inverted residuals and linear bottlenecks,'' in \emph{Proceedings of the IEEE conference on computer vision and pattern recognition}, 2018, pp. 4510--4520.

\bibitem{Gonzalez2009}
R.~C. Gonzalez, R.~E. Woods, and B.~R. Masters, ``Digital image processing, third edition,'' \emph{Journal of Biomedical Optics}, vol.~14, 2009.

\bibitem{Suzuki1985}
S.~Suzuki and K.~A. be, ``Topological structural analysis of digitized binary images by border following,'' \emph{Computer Vision, Graphics and Image Processing}, vol.~30, 1985.

\bibitem{Redmon2016}
J.~Redmon, S.~Divvala, R.~Girshick, and A.~Farhadi, ``You only look once: Unified, real-time object detection,'' vol. 2016-December, 2016.

\bibitem{Lhermitte2022}
E.~Lhermitte, M.~Hilal, R.~Furlong, V.~O’Brien, and A.~Humeau-Heurtier, ``Deep learning and entropy-based texture features for color image classification,'' \emph{Entropy}, vol.~24, 2022.

\bibitem{Freedman1981}
D.~Freedman and P.~Diaconis, ``On the histogram as a density estimator: L 2 theory,'' \emph{Z. Wahrscheinlichkeitstheorie verw. Gebiete}, vol.~57, pp. 453--476, 1981.

\bibitem{Nguyen2024}
V.~M. Nguyen, E.~Sandidge, T.~Mahendrakar, and R.~T. White, ``Characterizing satellite geometry via accelerated 3d gaussian splatting,'' \emph{Aerospace}, vol.~11, 2024.

\bibitem{Padilla2021}
R.~Padilla, W.~L. Passos, T.~L. Dias, S.~L. Netto, and E.~A.~D. Silva, ``A comparative analysis of object detection metrics with a companion open-source toolkit,'' \emph{Electronics (Switzerland)}, vol.~10, 2021.

\bibitem{Meni2024}
M.~Meni, T.~Mahendrakar, O.~D. Raney, R.~T. White, M.~L. Mayo, and K.~R. Pilkiewicz, ``Taking a peek into yolov5 for satellite component recognition via entropy-based visual explanations,'' 2024.

\bibitem{Selvaraju2020}
R.~R. Selvaraju, M.~Cogswell, A.~Das, R.~Vedantam, D.~Parikh, and D.~Batra, ``Grad-cam: Visual explanations from deep networks via gradient-based localization,'' \emph{International Journal of Computer Vision}, vol. 128, 2020.

\end{thebibliography}

\end{document}